\documentclass[journal,twoside,web]{ieeecolor}
\usepackage{tmi}
\usepackage{cite}
\usepackage{amsmath,amssymb,amsfonts}

\usepackage{multirow}
\usepackage{array} 
\usepackage{algorithmic}
\usepackage{graphicx}
\usepackage{textcomp}
\def\BibTeX{{\rm B\kern-.05em{\sc i\kern-.025em b}\kern-.08em
    T\kern-.1667em\lower.7ex\hbox{E}\kern-.125emX}}
\begin{document}
\title{MuVAM: A Multi-View Attention-based Model for Medical Visual Question Answering}
\author{Haiwei Pan, Shuning He, Kejia Zhang, Bo Qu, Chunling Chen, and Kun Shi 
\thanks{The work was supported by the International Exchange Program of Harbin Engineering University for Innovation-oriented Talents Cultivation, by the National Natural Science Foundation of China under Grant No.62072135 and No.61672181. (Corresponding author: Kejia Zhang.)
 }
\thanks{Haiwei Pan, Shuning He, Kejia Zhang, Chunling Chen, and Kun Shi are with the College of Computer Science and Technology, Harbin Engineering University, Harbin 150001, China (e-mail: panhaiwei@hrbeu.edu.cn; shuning@hrbeu.edu.cn; kejiazhang@hrbeu.edu.cn; ccl\_00@hrbeu.edu.cn; iamsxk@hrbeu.edu.cn).}
\thanks{Bo Qu is with Harbin Medical University, Harbin 150081, China (e-mail: heaven\_007@163.com).}
}

\maketitle

\begin{abstract}
Medical \emph{V}isual \emph{Q}uestion \emph{A}nswering (VQA) is a multi-modal challenging task widely considered by research communities of the computer vision and natural language processing. Since most current medical VQA models focus on visual content, ignoring the importance of text, this paper proposes a \emph{mu}lti-\emph{v}iew \emph{a}ttention-based \emph{m}odel(MuVAM) for medical visual question answering which integrates the high-level semantics of medical images on the basis of text description. Firstly, different methods are utilized to extract the features of the image and the question for the two modalities of vision and text. Secondly, this paper proposes a multi-view attention mechanism that include \emph{I}mage-to-\emph{Q}uestion (I2Q) attention and \emph{W}ord-to-\emph{T}ext (W2T) attention. Multi-view attention can correlate the question with image and word in order to better analyze the question and get an accurate answer. Thirdly, a composite loss is presented to predict the answer accurately after multi-modal feature fusion and improve the similarity between visual and textual cross-modal features. It consists of classification loss and \emph{i}mage-\emph{q}uestion \emph{c}omplementary (IQC) loss. Finally, for data errors and missing labels in the VQA-RAD dataset, we collaborate with medical experts to correct and complete this dataset and then construct an enhanced dataset, VQA-RAD$^{Ph}$. The experiments on these two datasets show that the effectiveness of MuVAM surpasses the state-of-the-art method. 
\end{abstract}

\begin{IEEEkeywords}
Attention Mechanism, Deep Learning, Medical Visual Question Answering, Multi-Modal Fusion, Medical Images.
\end{IEEEkeywords}

\section{Introduction}

\IEEEPARstart{A}{ccording} to different imaging principles, various levels of organ information is obtained, so as to understand our physical conditions comprehensively. Medical image \cite{chen2021super,al2019partial,jungo2018effect,tang2021disentangled,abdeltawab2020deep} plays an important role for doctors in clinical diagnosis. At the same time, electronic medical records of patients and diagnostic information provided by doctors are significant basis for disease analysis. Medical text also has a major impact on diagnosis \cite{li2021hybrid, fan2021deep, alfano2020design}. 
   
With the continuous development of intelligent medical care, computer-aided diagnosis technologies are gradually being recognized by the public. It uses medical image and text to help doctors make judgments on disease areas \cite{qiu2020novel,wang2018computer,ma2019computer}, thereby greatly reduce misdiagnosis and improve accuracy. Medical \emph{V}isual \emph{Q}uestion \emph{A}nswering (VQA)\cite{nguyen2019overcoming, zhan2020medical, vu2020question} is one of the hot topics in the current research of computer-aided diagnosis technology. It is a multi-modal challenging task that has been widely considered by the two main research directions of computer vision and natural language processing. In order to infer the correct answer, it is required to have a deep understanding of the rich content of medical images and precise exploration of the complex semantics of clinical questions. The purpose of this task is to assist the doctor in the diagnosis and alleviate the difficulties of patients in seeking medical treatment.

\begin{figure}[!t]
\centerline{\includegraphics[width=\columnwidth]{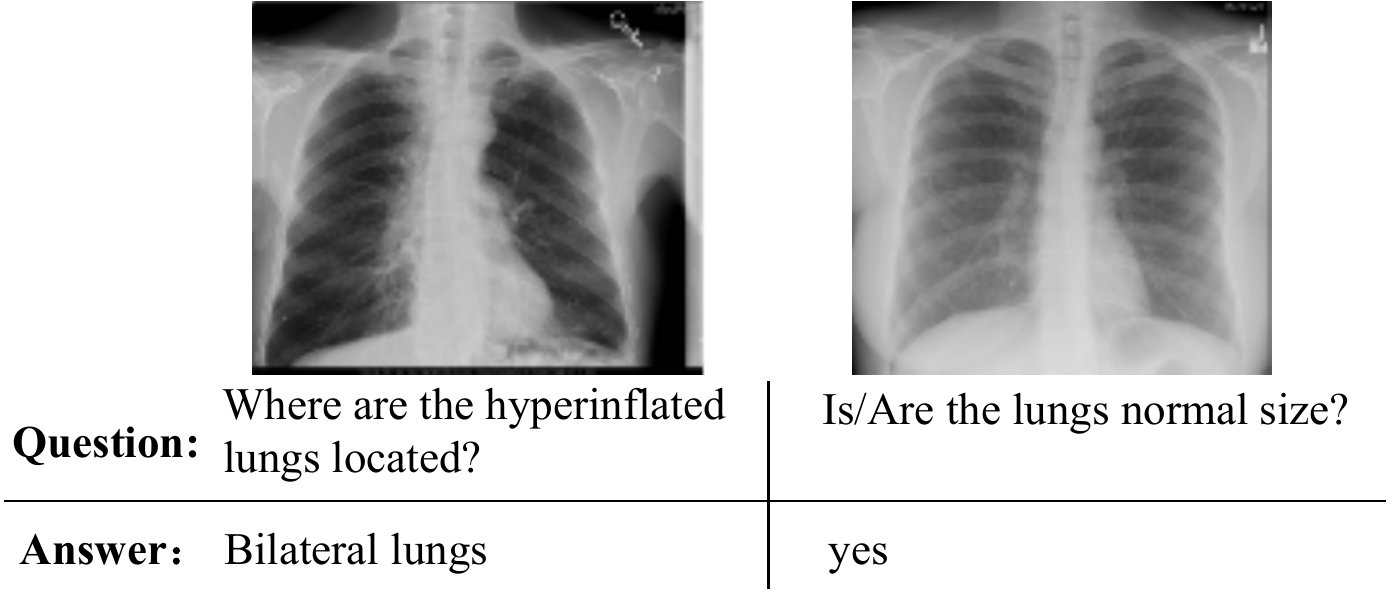}}
\caption{Using X-rays to take two extremely similar chest radiographs, the lung on the left is hyperinflated, and the lung on the right is of normal size.}
\label{fig1}
\end{figure}

However, the current research on VQA technology based on the professional medical fields is very limited. The imaging principle of medical images is complicated and the visual perception effect is not obvious. Most medical images of the same body part of different people are very similar, which is mainly caused by the high similarity of human structure. For example, the two extremely similar images are both chest radiographs taken through X-ray imaging, as shown in Fig.(\ref{fig1}). The bilateral lungs on the left image are abnormally hyperinflated, and the lungs on the right image are normal in size.

Although the deep learning method has been proven to be effective in medical image analysis \cite{xie2021survey}, the current medical VQA dataset lacks large-scale labeled training data. The labeling of medical images requires a lot of professional knowledge and time, which is not feasible on a large scale. If the deep learning model pre-trained on the general VQA dataset is transferred to medical VQA and fine-tuned with a small amount of medical images, the final effect is not satisfactory due to the obvious difference between medical images and natural images. Therefore, a medical VQA framework \cite{nguyen2019overcoming} overcomes the limitations of labeled data. \emph{M}odel-\emph{A}gnostic \emph{M}eta-\emph{L}earning (MAML) \cite{finn2017model} and \emph{C}onvolutional \emph{D}enoising \emph{A}uto-\emph{E}ncoder (CDAE) \cite{masci2011stacked} are used to initialize the weights of image feature extraction. MAML can quickly adapt to new tasks with a small number of training images.  The advantage of CDAE is to use large-scale unlabeled datasets and enhance the robustness of the model.  Although this method has achieved better effect which has done a lot of work on medical image feature extraction in the VQA-RAD dataset \cite{lau2018dataset}, VQA involves not only computer vision, but also text analysis of clinical questions.

 Due to the obscurity of professional medical concepts, there are also challenges in understanding clinical texts. A Question Conditional Reasoning (QCR) module is proposed by \cite{zhan2020medical}, which aims to extract task information from questions to guide the modulation of multi-modal features. In order to emphasize the important part of each question,  QCR applied an attention to obtain  importance weights  to different words, generally weakens the use of prepositions and articles, and highlights features such as nouns and verbs.

The above methods have a deep understanding of medical images and text questions respectively. However, independently modeling the semantic level of text and the visual  level of the image can not meet the requirements of multi-modal task. There are also correlations between images and questions. In this context, a \emph{mu}lti-\emph{v}iew \emph{a}ttention-based \emph{m}odel (MuVAM) for medical VQA is proposed which aims to fuse the high-level semantics of medical images on the basis of text. Firstly, the medical images and related questions are encoded and features are extracted. Then from the view of vision and words, \emph{i}mage-to-\emph{q}uestion (I2Q) attention and \emph{w}ord-to-\emph{t}ext (W2T) attention  are used  to act on the question. This process is called multi-view attention. In order to predict the answer accurately, a composite loss is proposed to train the MuVAM. It includes the classification loss after multi-modal fusion and \emph{i}mage-\emph{q}uestion \emph{c}omplementary (IQC) loss that combines image representation and text semantics to guide question importance learning.

\section{Related Works}
\subsection{Visual Question Answering}
Since the seminal work of \cite{antol2015vqa} was proposed, the task of VQA has attracted much research attention.The current VQA framework is mainly composed of a question feature extractor, an image feature extractor, and multi-modal fusion. The question feature extraction usually uses \emph{L}ong \emph{S}hort-\emph{T}erm \emph{M}emory(LSTM) \cite{hochreiter1997long}, \emph{G}ated \emph{R}ecurrent \emph{U}nits (GRU) \cite{cho2014learning}, and Skip-thought vectors \cite{kiros2015skip}. The mainstream image feature extraction method is to use Faster R-CNN \cite{ren2016faster} instead of the traditional CNN, so that the task is connected with the object detection to focus on the salient regions of the image related to the question \cite{teney2018tips}. However, in the professional field of medical VQA, this method is not applicable due to the lack of large-scale labeled training data in the dataset. In view of the particularity of the medical VQA dataset, MAML and CDAE were used to initialize the weights for image feature extraction in  \cite{nguyen2019overcoming}, so as to achieve the use of a small labeled training set to effectively train the effect of the medical VQA framework.

Early multimodal fusion generally used simple summation \cite{saito2017dualnet} , element-wise multiplication \cite{antol2015vqa} or vector concatenation \cite{zhou2015simple} methods to combine text and image information. With further research, bilinear pooling is applied to feature fusion gradually \cite{ben2017mutan, yu2017multi, yu2018beyond}.  It computes the outer product between two vectors for deep fusion. However, such operations lead to high feature dimensionality. Therefore, \emph{M}ultimodal \emph{C}ompact \emph{B}ilinear pooling (MCB) \cite{fukui2016multimodal} was proposed to perform the outer product calculation in a low-dimensional space to effectively combine multi-modal features. \emph{M}ultimodal \emph{L}ow-rank \emph{B}ilinear pooling (MLB) \cite{kim2016hadamard} was proposed to use Hadamard product and linear mapping to approximate the outer product to reduce computation complexity and reduce parameters. Bilinear pooling and attention mechanisms can also be combined. \emph{B}ilinear \emph{A}ttention \emph{N}etwork (BAN) \cite{kim2018bilinear} considered the bilinear interaction between the image and the question on the basis of low-rank bilinear pooling technique and proposed a variant of the multi-modal residual network to effectively utilize the multiple bilinear attention maps generated. Some studies \cite{li2018tell,wu2017image,wang2017vqa,li2017incorporating,wu2016ask} also extracted advanced semantic information from the attributes and visual relationships in the image. Image attributes  and  captions  of  the  model  mine  richer  VQA relationship in \cite{li2018tell}.

\subsection{Attention Mechanism}

\begin{figure*}[!t]
\centerline{\includegraphics[width=2\columnwidth]{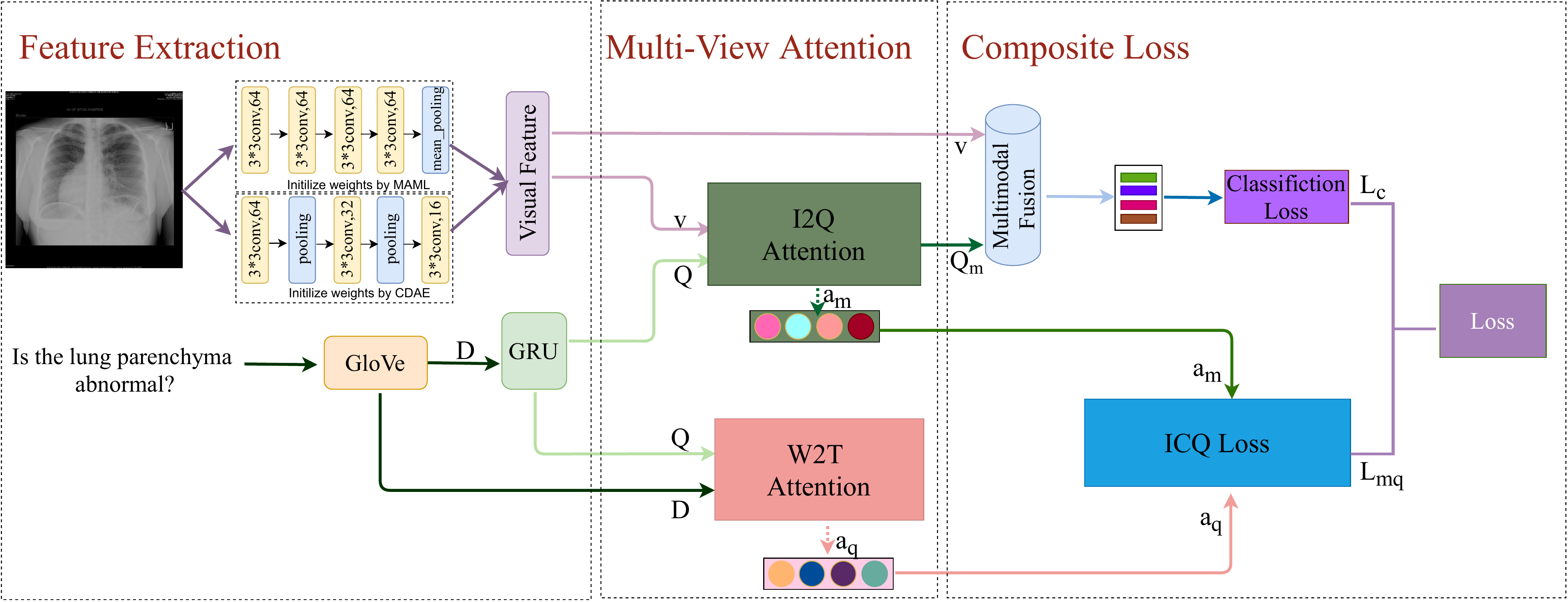}}
\caption{The training framework of MuVAM which includes a feature extraction module, a multi-view attention module and a compound loss module.}
\label{fig3}
\end{figure*} 

In order to obtain better results for VQA tasks, most of the methods currently are used to introduce the attention mechanism between feature extraction and multimodal fusion. When the proposed question is difficult to answer directly, a deeper understanding of the input question and the image is required. At this time, the attention mechanism appears particularly important. It turns out that the use of attention can improve the effect of most VQA tasks\cite{vu2020question}, and \cite{li2019relation}. The multi-glance attention mechanism was utilize in  \cite{vu2020question} to make the clinical question match the relevant regions in the medical image to get an accurate answer. In this way, a deeper meaning between the image and the question can be mined. The Relation-aware Graph Attention Network (ReGAT)  \cite{li2019relation} used the interactive dynamics between different objects to understand the visual scene in the image. The graph was composed of each image encoded. The semantic relations, spatial relations, and implicit relations between objects were modeled separately via a graph attention mechanism to learn the relation representations of adaptive questions. The attention weight of each object in the image was calculated to obtain question-oriented visual features that joint embedding of the question to feed to the classifier \cite{anderson2018bottom}. The re-attention framework\cite{guo2020re} associated images with questions from a fine-grained perspective by calculating the similarity of each object-word pairs in the feature space. Then, for the corresponding visual objects in the re-attending image, the initial attention map was reconstructed and the attention consistency loss was used to minimize the difference in the attention weight of the images.  A question-conditioned reasoning module to guide the selection of the importance of multi-modal fusion features was proposed \cite{zhan2020medical}. It used the attention mechanism of word embedding and question representations to obtain more semantic information to infer the answer.

Most attention mechanisms generally use the semantic representation of the question as a query to search for relevant image regions, and assign weights to the image to obtain a text-based image feature representation. In addition, the medical method lacks the attention of the image to the question. And the accuracies of the current methods need to be further improved. Therefore, we propose the MuVAM model to solve the above problems.

\section{MuVAM for Medical VQA}

The overall framework of MuVAM is shown in Fig. (\ref{fig3}). Given a medical image  $v$ and a question $Q$ related to the image, the correct answer is finally predicted. This can be specifically described as:

\begin{equation}
\hat{y}= \mathop{\arg\max}_{a\in\mathcal{A}}P(a\, | \, Q,v,\theta)
\end{equation}
where $\hat{y}$ represents the final classification result. $\mathcal{A}$ and $a$ respectively indicates a set of candidate answers and one of the answers. $P$ denotes the MuVAM framework and $\theta$  represents all the parameters of the  MuVAM. 
The  MuVAM framework is divided into three parts, which are feature extraction module, multi-view attention module, and composite loss module. In the feature extraction module, different methods are used to extract features for the medical image and the related question. Then the image feature and the question feature are fed to the multi-view attention module, which includes W2T attention and I2Q attention. They pay attention to the question from the views of words and images to obtain text features under the guidance of vision and two attention weights. Finally, the output of the multi-view attention module is passed to the composite loss module. In order to maximize the accuracy of the prediction results, the composite loss is composed of the classification loss and the IQC loss to train MuVAM together. Classification loss predicts answer distribution and IQC loss minimizes the difference between the importances of question learned by words and learned under the guidance of vision. The three modules are explained in detail in the following subsections.

\subsection{Feature Extraction}

Because different modalities have completely different description forms and complex coupling correspondences, it is necessary to solve the multi-modal data representation in a unified way. The accuracy of the extracted features is critical to the subsequent operations. In this part, different feature extraction methods are used for the image and the question respectively. The information of the two modalities is mapped to the same feature space to obtain feature representations.

\subsubsection{Image representation}

At present, most image feature extraction is based on the ResNet CNN within a Faster R-CNN framework \cite{ren2016faster} to extract more fine-grained image representations. Because this method requires a large amount of labeled training data that is not suitable for medical VQA problem for which the available datasets are relatively small, this paper adopts MAML and CDAE encoder to initialize the pre-trained weights for image representation \cite{nguyen2019overcoming}. MAML consists of four 3*3 conolutional layers with stride 2 and ends with a mean pooling layer.Each convolutional layer contains 64 filters and a ReLu layer. CDAE encoder is a combination of a stack of convolutional layers and max pooling layers. MAML and CDAE encoder obtain 64-dimensional vector features respectively, which are concatenated to generate 128-dimensional  image features. The feature of the image is represented as $ v\in \mathcal{R}^{d_k} $, $d_{k} = 128$ denotes the dimension of image features. 

\subsubsection{Question representation}

Each question is unified into a sentence consisting of $n$ words. Parts exceeding $n$ words will be deleted and parts less than $n$ words will be zero-padded . Words are represented by 300-dimensional GloVe word embedding  $D =\{w_1, w_2...w_n \} \in \mathcal{R}^{d_{h}*n}$, $d_{h}=300$ denotes the dimension of each word representation. Finally, the word vectors are fed to the Gated Recurrent Units  (GRU) \cite{cho2014learning} network to encode the question embedding $Q =\{q_1, q_2...q_n \} \in \mathcal{R}^{d_{s}*n}$, $d_{s}=1024$ is the dimension of each hidden state in GRU.

\subsection{Multi-View Attention }

For a question-answer pair, there is a correlation between the image and the question. In order to maximize the application of semantic information, a multi-view attention mechanism is proposed. It includes word-to-text attention and image-to-question attention, giving attentions to the question from two views of the word and the image.

\subsubsection{W2T attention mechanism}

We use the question conditional reasoning framework proposed by \cite{zhan2020medical} to emphasize the important part of the question. The question representation $Q$ obtained by the feature extraction module has the same importance for different words, which is inconsistent with human brain focusing. In order to emphasize the significant part of the question, the attention mechanism is used to give full play to the advantages of word embedding and sentence vectors to assign different words with importance weights. This part gets the importance of each word in the question from the semantic level. Firstiy, the word embedding representation $D$ and the question embedding representation $Q$ are concatenated to obtain $Q_{c}$.

\begin{equation}
Q_{c}= [D \,|| \,Q]
\end{equation}
where $||$ represents the concatenation of feature dimensions, $Q_{c} \in R^{(d_h+d_s)*n}$. 

Then the sigmoid activation function is used as a selection mechanism to filter out useless noise to control the output. We take the advantages of context-free (GloVe) and contextual embedding (GRU)  to obtain $\widetilde Q \in R^{d_s*n}$.

\begin{equation}
\widetilde Q =  \tanh(W_{1}Q_{c}) \odot \sigma(W_{2}Q_{c})
\end{equation}
where $W_1, W_2\in R^{d_s*(d_h+d_s)}$ are learned weights, $\odot$ is the Hadamard product, $\tanh(.)$ and $\sigma(.)$ are activation functions, called sigmoid and gated hyperbolic tangent respectively. 

 Finally, we get the attention weight $a_q\in R^{n*1} $ on the semantic level for the question embedding $Q$.

\begin{equation}
a_q =  softmax((W_{3}\widetilde Q)^T)
\end{equation}
where $ W_3\in R^{1*d_s}$ is learned weights.

\subsubsection{I2Q attention mechanism}

In this part, we propose an I2Q attention mechanism that is introduced to establish the relation between the two modalities to observe the question from a visual perspective. This mechanism is derived from the Rosenthal effect in social psychology. It refers to a phenomenon that the ardent hopes of teachers for students can achieve the expected results dramatically. Inspired by the Rosenthal effect, we designed the Rosenthal expectation to correspond the image and the question to the teacher and the student respectively.  The image assigns different attention weights to the words in the question and mines the most potential words. Based on this idea, the attention weight is used to accurately mine the degree of strong association between the image and the words in the question.

\begin{equation}
\label{eq5}
a_m =  softmax(Q^{T}MLP(v))
\end{equation}
where $MLP(.) $ is a multi-layer perceptron used to align the dimension between $Q$ and $v$.  Eq. \eqref{eq5} can be used to obtain the weight distribution $a_m\in R^{n*1} $ of the $n$ keywords assigned to the question by the image in the question-answer pair. Each element in $a_m$ corresponds to the degree of correlation between the keyword and the image. The larger the value of the element, the higher the correlation.

After generating the attention weight matrix $a_m$ of the text guided by the vision, the result of Eq. \eqref{eq5}  is applied to the question embedding $Q$ of feature extraction. Finally, the module obtains the question embedding $Q_{m}$ with the image features fused. At this time, the question embedding not only includes the relevant features of the semantic mono-modality of the text, but also adds the image features.  The features of these two modalities are combined through the I2Q attention. The combined features give full play to their respective advantages to achieve the purpose of accurately judging the fine-grained relationship between text and vision. According to the degree of relevance of each keyword to the image, different weights are assigned to each word in the question. The feature of the question affected by the vision can be obtained as shown in the following:

\begin{equation}
{Q_m}= {a_m}^T \odot Q
\end{equation}
where $\odot$ is the Hadamard product and finally the text feature ${Q_m}\in R^{d_s*n}$ learned under the guidance of vision is obtained.

\subsection{Composite Loss}

In the process of model optimization, in order to maximize the accuracy of VQA, we design a composite loss function with the following loss conditions to train the entire model. The loss module contains a total of two parts, namely classification loss and image-question complementary (IQC) loss.

\subsubsection{Classification Loss}

 After obtaining the text features of the visual guidance, we use the TCR module proposed by \cite{zhan2020medical} to divide question-answer pairs into open-end and closed-end based on answer types  in order to compare the accuracy of different types of question-answer pairs. The question representation ${Q_m}$ and the image feature $v$ of the two types are respectively passed into the general multimodal fusion model, and the fused multimodal features are output:

\begin{equation}
M^{cl}= \mathcal{F}_{\Theta}(v^{cl}, {Q_m^{cl}})
\end{equation}

\begin{equation}
M^{op}= \mathcal{F}_{\Theta}(v^{op}, {Q_m^{op}})
\end{equation}
where $\mathcal{F}$ is a multi-modal feature fusion representation method. $cl$ and $op$ stand for closed-ended and open-ended respectively. In this paper, bilinear attention network (BAN)  \cite{kim2018bilinear} is used to learn joint representation, and $\Theta$ is a trainable parameter in the fusion process. 

The multi-modal features $M^{cl}$ and $M^{op}$ are respectively passed to the classifier to predict the best answer. The classifier is composed of two-layer MLP to obtain the probabilities of candidate answers. The answers with the maximum probability in all candidate answers are selected as the final prediction $\hat {y^{cl}}$ and $\hat {y^{op}}$. In this stage, the binary cross-entropy loss $L_{c}$ is used in the training process.

\begin{equation}
L_{c}= BCE(\hat{y^{cl}} , y^{cl}) + BCE(\hat{y^{op}} , y^{op})
\end{equation}
where $BCE(.)$ represents the binary cross-entropy loss function, $\hat{y}$ is the predicted answer and $y$ is the ground-truth answer. $cl$ and $op$ stand for closed-ended and open-ended respectively.

\subsubsection{IQC loss}

This part uses the learned weight $a_m$ obtained by the W2T attention and the attention weight $a_q$ generated by the I2Q attention to jointly guide the learning of question importance.

Generally Multi-modal tasks use the complementarity property between multiple modalities to not only eliminate the redundancy, but also obtain more comprehensive and accurate information to enhance the reliability and fault tolerance of task learning. In the model training process,  this paper minimizes the difference between the importances of question learned by words and learned under the guidance of vision in order to improve the similarity between visual-text cross-modal features. This method is to use ${a_m}$ and ${a_q}$ to define the image-question complementary loss$L_{mq}$:

\begin{equation}
L_{mq} =  \left\|a_m - a_q\right\|_2^2
\end{equation}

Therefore, the two parts of the composite loss module are used to jointly optimize the MuVAM:

\begin{equation}
\label{eq10}
Loss = L_c + \gamma L_{mq}
\end{equation}
where  $\gamma$ are hyper-parameters, and the final value is estimated in our experiment. Facts have proved that the IQC loss  is effective in improving the accuracy of our model.

\begin{figure*}[!t]
\centerline{\includegraphics[width=2\columnwidth]{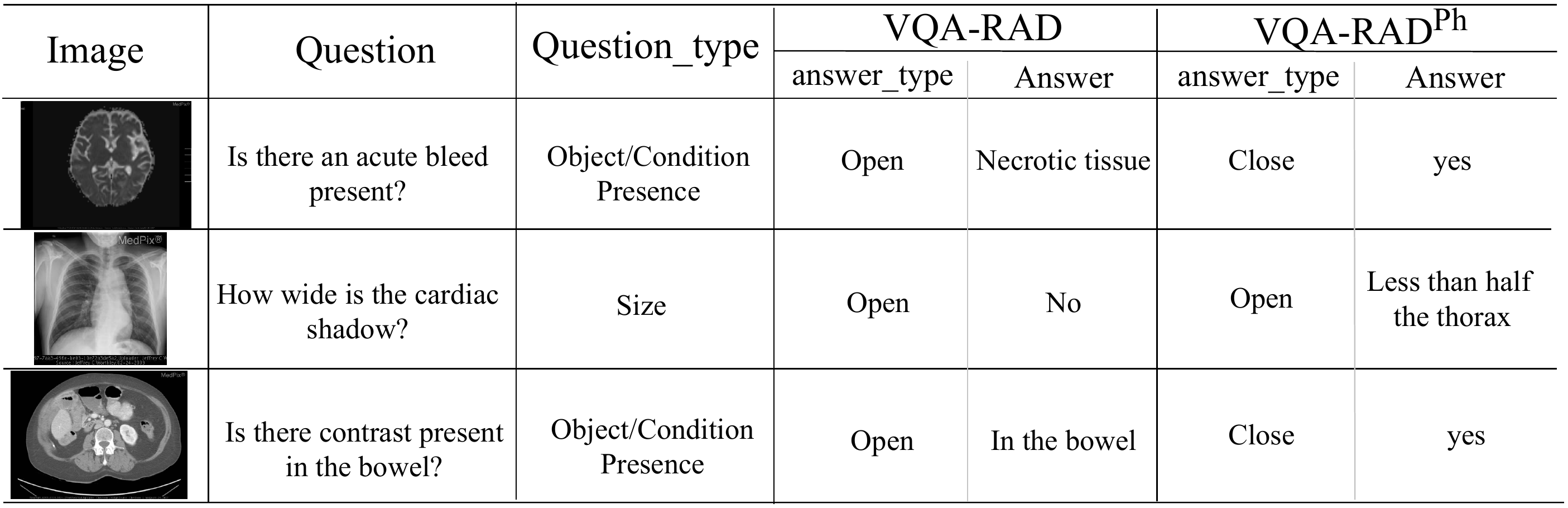}}
\caption{Examples of VQA-RAD dataset and VQA-RAD$^{Ph}$ dataset.}
\label{fig4}
\end{figure*} 

\begin{figure}[!t]
\centerline{\includegraphics[width=\columnwidth]{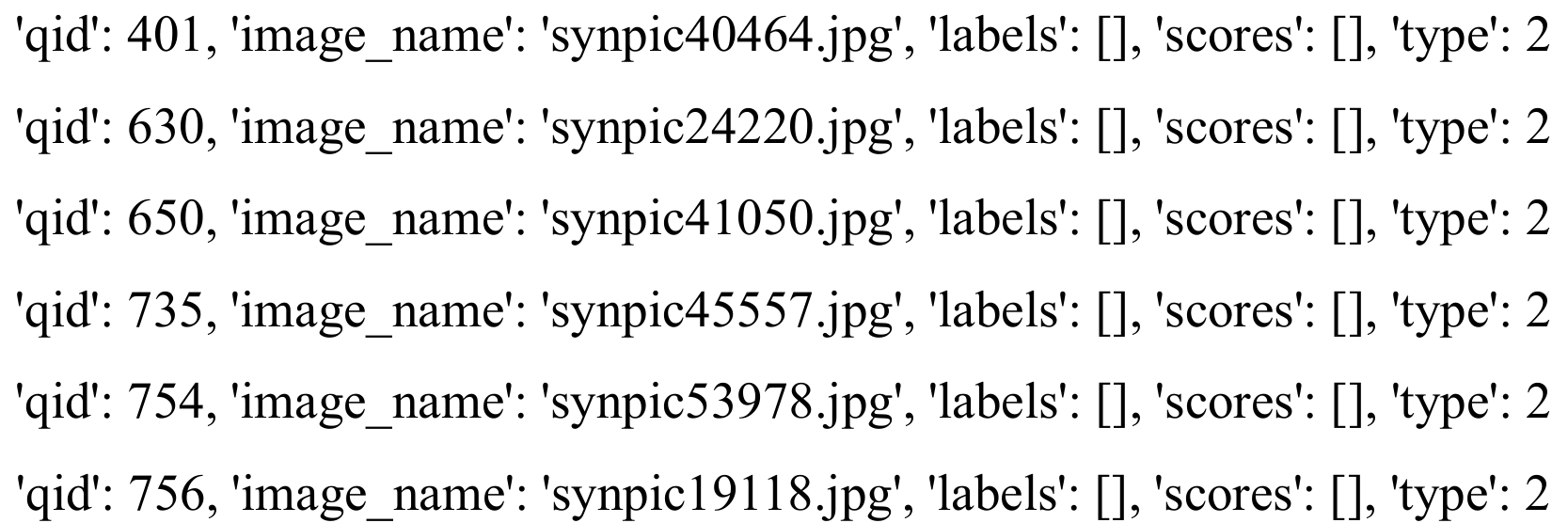}}
\caption{The anomaly data of VQA-RAD dataset.}
\label{fig7}
\end{figure} 

\section{Experiment}

In this section, we conduct comprehensive experiments to evaluate the effectiveness of MuVAM. Firstly, we collaborate with medical experts to correct and complete the VQA-RAD datasetand then construct an enhanced dataset, VQA-RAD$^{Ph}$. Secondly, MuVAM is compared with other current advanced models \cite{nguyen2019overcoming,zhan2020medical}, and \cite{do2021multiple} on the VQA-RAD dataset. And MuVAM and Med-VQA models are evaluated on VQA-RAD$^{Ph}$. Next, The ablation study is used to prove the necessity of the components and the hyperparameter $\gamma$  of the image-question complementary loss (IQC) is given the best value. Finally, the visualization evaluation is used to qualitatively analyze the MuVAM model.

\subsection{Datasets and data quality improvement}

\subsubsection{VQA-RAD}

The VQA-RAD dataset is the first manually constructed dataset in the field of medical VQA. It contains a total of 315 medical radiology images that are evenly distributed on the head, chest and abdomen. The dataset is divided into a training set and a test set which contains 3064 and 451 question-answer pairs. Questions are classified into 11 categories, for example, organ system, modality, and location.  According to the answer types, The question-answer pairs are split into open-ended and close-ended. Generally, the answer that is "Yes/no" or to the either-or choice question is defined as the close-ended. The majority of the organ system and positional questions answered in the form of free text are open-ended.  In our model, we use the TCR component proposed by \cite{zhan2020medical} to distinguish these two types.

\subsubsection{VQA-RAD$^{Ph}$}

The most common application of VQA in reality is to help the blind and visually impaired to obtain more information on the Internet or in the real world. It can even carry out real-time human-computer interaction, which will greatly improve their living conditions and convenience. But if there is a problem in the dataset itself and the wrong answer is eventually returned, then the incorrect answer in this situation will lead to a fatal accident. Especially for the VQA dataset in the medical vertical field, data quality is more emphasized.We invited medical imaging experts to check the dataset one by one and it was found that the question-answer pairs in the VQA-RAD dataset are not completely correct. As an example is shown in Fig. (\ref{fig4}), for the question "Is there an acute bleed present?", this is obviously a closed-ended question, and the correct answer is "Yes". However, the reference answer in VQA-RAD dataset is the"Necrotic tissue" which is open-ended. We screened out the questionable data from all the question-answer pairs, discussed and revised them with experts in multiple medical fields to correct and complete the dataset. 

In the experiment, it was found that a total of 48 question-answer pairs in the test set were not labeled is shown in Fig. (\ref{fig7}). This phenomenon is called the fifth anomalous task in \cite{lee2020regularizing}, that is, the undefined answer. Although the undefined answer means that the image and question are from in-distribution, the answer is not among the candidate answers of the VQA model. When the correct answer does not exist, there will be an unanswerable situation. This situation is also the most common and realistic anomaly. In order to improve data quality and make the dataset available , we repair the anomaly data, and finally expand from 486 different answers to 521 according to the unique structure of the answer. The correct and complete dataset is called VQA-RAD$^{Ph}$.

\subsection{Implementation Setup}

All experiments are implemented on a server with GTX 1080ti GPU using pytorch. Each question is composed of 12 words and each word is embedded with GloVe. The sequence of word embedding is fed into GRU to obtain semantic features. The dimension of the hidden layer in GRU is set to 1024. The visual feature extraction component is composed of MAML and CDAE encoder  to initialize the pre-training weights and the size of the relation feature is set to 128.  BAN \cite{kim2018bilinear} is used as a multi-modal feature fusion model in the fusion module.  In our experiment, we use the Adamax optimizer for training with the minimum batch size as 64 and the learning rate is set to 0.005.

\subsection{Evaluation Metrics}

The accuracy is used as the evaluation index of the model in the experiment, that is, the proportion of the total amount that the model predicts correctly:

\begin{equation}
P_{Acc} = \frac{N_C}{N_T}*100\%
\end{equation}
where $N_C$ represents the number of correctly predicted answers to the questions, $N_T$ refers to the total number of questions.

\subsection{Baseline}

We compare the proposed model with the advanced methods of medical VQA: MAML \cite{finn2017model}, MEVF \cite{nguyen2019overcoming}, MMQ \cite{do2021multiple} , Med-VQA \cite{zhan2020medical}.

 \textbf{MAML} \cite{finn2017model} introduced model-agnostic meta-learning to learn meta-weights that quickly adapt to visual concepts to overcome the problems caused by transfer learning to achieve image feature extraction.
 
 \textbf{MEVF} \cite{nguyen2019overcoming} leveraged meta-learning and denoising auto-encoder to extract image features in order to overcome the limitation of labeled training data.
 
 \textbf{MMQ} \cite{do2021multiple}  aimed to increase metadata through automatic annotations and process noisy labels , without external data to train the meta-model for better feature extraction of medical images. 
 
 \textbf{Med-VQA} \cite{zhan2020medical} proposed a conditional reasoning framework that included QCR component and TCR component, aiming to automatically learn effective reasoning skills for various medical VQA tasks.
 
 MAML, MEVF and MMQ use SAN \cite{yang2016stacked} and BAN \cite{kim2018bilinear} fusion methods. Med-VQA and MuVAM use BAN fusion method.

\begin{table}[h]
\renewcommand\arraystretch{1.3}
\caption{Comparison of the accuracy of different medical vision question answering methods in the VQA-RAD test set. }
\label{table1}
\setlength{\tabcolsep}{3.5mm}{
\begin{tabular}{c|c|c|c|c}
\hline
\multicolumn{2}{c|}{\multirow{3}{*}{Model}} & \multicolumn{3}{c}{VQA-RAD}                                                                                                                                                    \\ \cline{3-5} 
\multicolumn{2}{c|}{}                       & \begin{tabular}[c]{@{}c@{}}Open-ended\\ (\%)\end{tabular} & \begin{tabular}[c]{@{}c@{}}Closed-ended\\ (\%)\end{tabular} & \begin{tabular}[c]{@{}c@{}}Overall\\ (\%)\end{tabular} \\ \hline
\multirow{2}{*}{MAML}       & BAN      & 40.1                                                      & 72.4                                                       & 59.6                                                   \\ \cline{2-5} 
                                  & SAN      & 38.2                                                      & 69.7                                                       & 57.1                                                   \\ \hline
\multirow{2}{*}{MEVF}       & BAN      & 43.9                                                      & 75.1                                                       & 62.7                                                   \\ \cline{2-5} 
                                  & SAN      & 40.7                                                      & 74.1                                                       & 60.7                                                   \\ \hline
\multirow{2}{*}{MMQ}        & BAN      & 53.7                                                      & 75.8                                                       & 67.0                                                   \\ \cline{2-5} 
                                  & SAN      & 46.3                                                      & 75.7                                                       & 64.0                                                   \\ \hline
\multicolumn{2}{c|}{Med-VQA}          & 60.0                                                      & 79.3                                                       & 71.6                                                   \\ \hline
\multicolumn{2}{c|}{\textbf{MuVAM}}          & \textbf{63.3}                                             & \textbf{81.1}                                              & \textbf{72.2}                                                   \\ \hline
\end{tabular}}
\end{table}

\begin{table}[h]
\renewcommand\arraystretch{1.2}
\caption{ Comparison of the accuracy of different medical vision question answering methods in the VQA-RAD$^{Ph}$ test set.}
\label{table2}
\setlength{\tabcolsep}{4.5mm}{
\begin{tabular}{c|c|c|c}
\hline
\multirow{3}{*}{Model} & \multicolumn{3}{c}{VQA-RAD$^{Ph}$}                                                                                                                                                    \\ \cline{2-4} 
                       & \begin{tabular}[c]{@{}c@{}}Open-ended\\ (\%)\end{tabular} & \begin{tabular}[c]{@{}c@{}}Closed-ended\\ (\%)\end{tabular} & \begin{tabular}[c]{@{}c@{}}Overall\\ (\%)\end{tabular} \\ \hline
Med-VQA                &56.11                                                          & 78.96                                                           & 68.07                                                      \\ \hline
\textbf{MuVAM}                  & \textbf{63.33}                                                       & \textbf{82.66}                                                       & \textbf{74.28}                                                  \\ \hline
\end{tabular}}
\end{table}

\begin{table*}[h]
\renewcommand\arraystretch{1.4}
\caption{Analysis of the VQA-RAD and VQA-RAD$^{Ph}$ testset used for ablation research, where “baseline” is the basic model,  the "att" refers to the Image-to-Question attention mechanism , and the "$L_{mq}$" indicates image-question complementary loss. }
\label{table3}
\setlength{\tabcolsep}{6mm}{
\begin{tabular}{c|ccc|ccc}
\hline
\multirow{2}{*}{Method} & \multicolumn{3}{c|}{VQA-RAD}                                                                                                                                                     & \multicolumn{3}{c}{VQA-RAD$^{Ph}$}                                                                                                                                                                                                                                                                                                     \\ \cline{2-7}
                        & \begin{tabular}[c]{@{}c@{}}Open-ended\\ (\%)\end{tabular} & \begin{tabular}[c]{@{}c@{}}Closed-ended\\ (\%)\end{tabular} & \begin{tabular}[c]{@{}c@{}}Overall\\ (\%)\end{tabular} & \begin{tabular}[c]{@{}c@{}}Open-ended\\ (\%)\end{tabular} & \begin{tabular}[c]{@{}c@{}}Closed-ended\\ (\%)\end{tabular} & \begin{tabular}[c]{@{}c@{}}Overall\\ (\%)\end{tabular} \\ \hline
baseline                & 57.01                                                     & 78.79                                                       & 70.06                                                  & 57.22                                                     & 79.11                                                       & 70.28                                                  \\
baseline+att            & 60.00                                                     & 81.54                                                       & 72.06                                                  & 62.77                                                     & 80.07                                                       & 72.50                                                  \\ 
baseline+$L_{mq}$            & 61.66                                                     & 81.48                                                      & 71.62                                                  & 59.44                                                     & 81.18                                                       & 71.61                                                  \\ 
baseline+att+$L_{mq}$        & 63.33                                                    & 81.11                                                       & 72.28                                                  & 63.33                                                     & 82.66                                                       & 74.28                                                  \\ \hline
\end{tabular}}
\end{table*}

\begin{table*}[h]
\renewcommand\arraystretch{1.5}
\caption{$ \gamma $changes from 0 to 2.0 when the interval is 0.2 in Eq. \eqref{eq10} . The accuracy of the proposed model is evaluated in the VQA-RAD$^{Ph}$ testset. }
\label{table4}
\setlength{\tabcolsep}{3.3mm}{
\begin{tabular}{|c|c|c|c|c|c|c|c|c|c|c|c|c|}
\hline
\multirow{2}{*}{Method}& \multirow{2}{*}{Type} & \multicolumn{11}{c|}{$\gamma$}                                                                       \\ \cline{3-13} 
                        &                       & 0     & 0.2   & 0.4   & 0.6   & 0.8   & 1.0   & 1.2   & 1.4   & \textbf{1.6}            & 1.8   & 2.0   \\ \hline
\multirow{3}{*}{MuVAM}   & Open-ended            & 62.77 & 61.67 & 62.22 & 60.00 & 61.11 & 62.22 & 60.56 & 61.67 & \textbf{63.33} & 62.22 & 60.77 \\ \cline{2-13} 
                        & Closed-ended          & 80.07 & 80.81 & 80.81 & 80.00 & 80.40 & 81.18 & 81.92 & 79.70 & \textbf{82.66} & 80.00 & 80.44 \\ \cline{2-13} 
                        & Overall               & 72.50 & 72.28 & 72.06 & 71.62 & 71.62 & 72.51 & 72.51 & 70.95 & \textbf{74.28} & 72.28 & 71.39 \\ \hline
\end{tabular}}
\end{table*}

The results of our model and other advanced methods using the VQA-RAD dataset are compared in TABLE \ref{table1}. It can be shown that our results are superior to other baseline models in open-ended, closed-ended questions and overall. Compared with the Med-VQA method, all accuracy rates are improved by about 3\% on average. MuVAM models the relationship between text and vision. Benefiting from it, the questions and images can be better understood to generate satisfactory answers. This shows the importance of the association between each word and the image, thus demonstrats the effectiveness of the MuVAM.

We use the VQA-RAD$^{Ph}$ dataset to evaluate the Med-VQA \cite{zhan2020medical} and the MuVAM proposed in this paper. The results in TABLE \ref{table2} show that the MuVAM is more effective. The experimental effect of Med-VQA in the VQA-RAD$^{Ph}$ dataset is slightly inferior to the test results in the VQA-RAD dataset. The performance of MuVAM on open-ended questions in these two datasets is basically the same. The closed-ended questions and overall results are more prominent in the VQA-RAD$^{Ph}$ dataset.

\begin{figure*}[!t]
\centerline{\includegraphics[width=2\columnwidth]{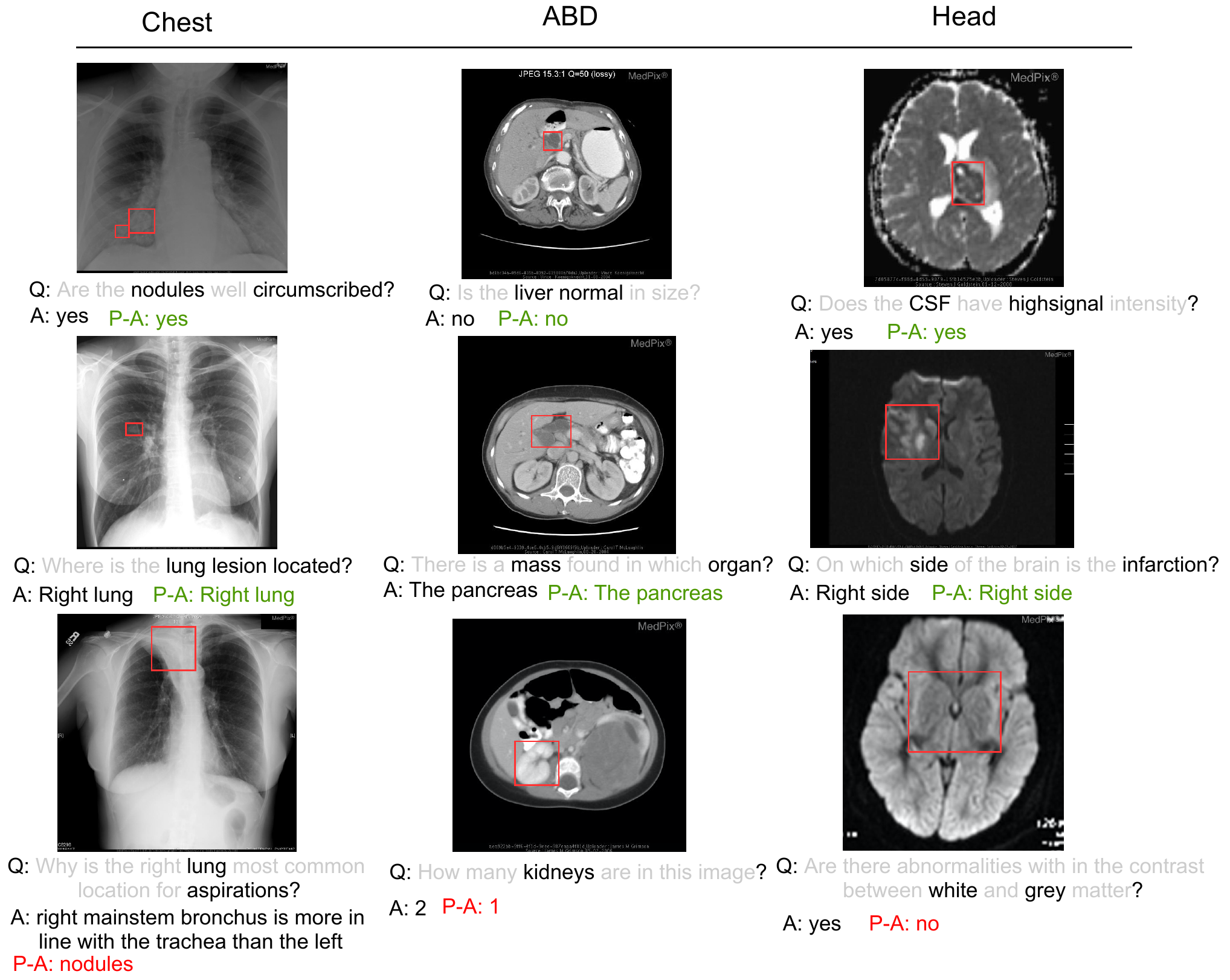}}
\caption{The visualization of the proposed model, each typical example includes image, question, ground-truth answer, and predicted answer. The rectangular box in the image represents the relevant region located by the model, and the black text of the question denotes the keywords involved. A refers to the ground-truth answer. The green and red content of P-A indicate that the answer predicted by the model is correct and wrong.}
\label{fig5}
\end{figure*} 

\subsection{Ablation Study}

In this section, complete ablation experiments was implemented to verify the performance of the I2Q attention and the IQC loss in MuVAM on the VQA-RAD$^{Ph}$ dataset and VQA-RAD dataset. The basic model is "MEVF+BAN+TCR" with only classification loss in the loss module, which is called the baseline, where "TCR" is a component from \cite{zhan2020medical} and "MEVF+BAN" is the method of \cite{nguyen2019overcoming}. For fairness, the baseline is re-implemented for the following ablation studies when the experimental parameters and equipment are consistent with our experiments. 

The experimental results of the ablation study are shown in TABLE \ref{table3}. The proposed components are evaluated on VQA-RAD and VQA-RAD$^{Ph}$ respectively. "Baseline+att" is used to verify the I2Q attention. "Baseline+$L_{mq}$" represents a benchmark model with image-to-question complementary loss and the purpose is to improve the similarity between visual-text cross-modal features. "Baseline+att+$L_{mq}$" implies that the attention and loss items work together. It can be seen from the results that the cooperation  between the two components is superior to that of any one of them, but those are better than the baseline. Horizontal comparison shows that the volatility of "baseline+att" and "baseline+$L_{mq}$" is relatively large in the open-ended questions of the two datasets. However, when "att" and "$L_{mq}$" work together, they have a stabilizing effect and the best results. Most of the experimental results of VQA-RAD$^{Ph}$ are also better than VQA-RAD, indicating that the VQA-RAD$^{Ph}$ dataset improves the data quality  and enhances the data availability.

\subsection{Hyperparameter Analysis}

In this section, we assign different values to the hyperparameters $\gamma$ of image-question complementary loss in Eq. \eqref{eq10} to analyze the performance of our proposed method. When different values of $\gamma$ are set in three different types: Open-ended, Closed-ended and Overall, the detailed accuracies of the proposed model are verified in TABLE \ref{table4}.

The results show that the performance varies along with the $\gamma$. As shown in TABLE \ref{table4}, when $\gamma$ is 1.6, the accuracies of the three types are particularly prominent. The best results that can be obtained are 63.33\%, 82.66\% and 74.28\%, respectively. Compared with the accuracies when $\gamma$ is 0, the three types have been improved, so it can prove again that the IQC loss has a significant effect on the MuVAM model. In the section of ablation study and baseline, we set $\gamma$ to 1.6.

\subsection{Visualization Evaluation}

the visualization evaluation of the MuVAM model on the VQA-RAD$^{Ph}$ dataset is shown in Fig. \eqref{fig5}. The rectangular box of the image is the key part of the positioning and the bold words in the question are the meaningful words learned by the attention mechanism. P-A represents the answer predicted by the method, where green indicates correct prediction, and red indicates incorrect prediction. $A$ is the ground-truth answer. MuVAM can find the visual information and the key words of the text involved in the VQA tasks accurately. Taking the first closed-ended question of the chest as an example, our method gives the correct answer to this question according to location of  the relevant region and the keywords learned through the attention mechanism focus on ”nodule” . It is worth noting that the actual location of the radiograph image is the exact opposite of what we see. The second sample of the head, the ground-truth answer "right side" is predicted by an accurate understanding of the question and the image.

  However, our model cannot always predict correctly. On the one hand, some common sense and experience are involved. taking the third sample of the chest as an example, the questions need to be answered based on the experience of professional doctors.  It is impossible to answer the questions based on the attention mechanism alone. The proposed model cannot satisfy the prediction of such related tasks. On the other hand, the model fails to locate keywords and visual content accurately. For example, in the third task of the head, the model does not pay attention to the keyword "adnormalities" . This makes it impossible to predict the correct answer in the end. These observations can help us make subsequent improvements to the model.

\section{Conclusion}

In this paper, MuVAM is proposed which can solve medical VQA tasks effectively. It contains three modules. First, two feature extraction methods are used in the feature extraction module to obtain image representation and question representation. Second, in order to maximize the application of semantic information, this paper proposes a multi-view attention module. It includes image-to-question (I2Q) attention and word-to-text (W2T) attention, which explore the potential impact on the question from the word and the image. Third, a compound loss module is proposed to train the model to improve the accuracy of MuVAM. It consists of classification loss and image-question complementary (IQC) loss. It is worth noting that the IQC loss uses image representation and text semantics to  jointly guide the question importance learning to strengthen the role of similarity and weaken the difference in visual-text cross-modal features. Our experiment corrects and completes the VQA-RAD dataset and constructs an enhanced dataset called VQA-RAD$^{Ph}$ to improve data quality. Various ablation studies are demonstrated the effectiveness of the proposed components for medical VQA tasks. And our method outperforms the state-of-the-art method.

\end{document}